\documentclass[a4paper,twoside]{article}

\usepackage{epsfig}
\usepackage{subcaption}
\usepackage{calc}
\usepackage{amssymb}
\usepackage{amstext}
\usepackage{amsmath}
\usepackage{amsthm}
\usepackage{multicol}
\usepackage{pslatex}
\usepackage{apalike}
\usepackage{algorithm2e}
\usepackage[bottom]{footmisc}
\usepackage{graphicx}
\usepackage{amsmath}
\usepackage{booktabs}

\usepackage{times}
\usepackage{multirow}
\usepackage{rotating}
\usepackage{pifont}

\usepackage{url}
\usepackage{siunitx}

\usepackage{slashbox}
\usepackage[hidelinks]{hyperref}
\usepackage[sort&compress,capitalize,noabbrev,nameinlink]{cleveref}
\usepackage[acronym]{glossaries}

\usepackage{SCITEPRESS}     

\newcommand\Tstrut{\rule{0pt}{2.4ex}}         

\creflabelformat{equation}{#2#1#3}

\newacronym{snn}{SNN}{Spiking Neural Network}
\newacronym{ann}{ANN}{Artifical Neural Network}
\newacronym{dl}{DL}{Deep Learning}
\newacronym{lif}{LIF}{Leaky Integrate-and-Fire}
\newacronym{if}{IF}{Integrate-and-Fire}
\newacronym{stdp}{STDP}{Spike Timing Dependent Plasticity}
\newacronym{dvs}{DVS}{Dynamic Vision Sensor}
\newacronym{bptt}{BPTT}{Back-Propagation Through Time}
\newacronym{sgd}{SGD}{Stochastic Gradient Descent}

\begin{document}

\title{ShapeAug: Occlusion Augmentation for Event Camera Data}

\author{\authorname{Katharina Bendig\sup{1},  René Schuster\sup{1,2}, Didier Stricker\sup{1,2}}
\affiliation{\sup{1}RPTU -- University of Kaiserslautern-Landau}
\affiliation{\sup{2}DFKI -- German Research Center for Artificial Intelligence}
\email{firstname.lastname@dfki.de}
}


\abstract{Recently, \Glspl*{dvs} sparked a lot of interest due to their inherent advantages over conventional RGB cameras. These advantages include a low latency, a high dynamic range and a low energy consumption. Nevertheless, the processing of \gls{dvs} data using \gls{dl} methods remains a challenge, particularly since the availability of event training data is still limited. This leads to a need for event data augmentation techniques in order to improve accuracy as well as to avoid over-fitting on the training data. Another challenge especially in real world automotive applications is occlusion, meaning one object is hindering the view onto the object behind it. In this paper, we present a novel event data augmentation approach, which addresses this problem by introducing synthetic events for randomly moving objects in a scene. We test our method on multiple \gls{dvs} classification datasets, resulting in an relative improvement of up to 6.5 \% in top1-accuracy. Moreover, we apply our augmentation technique on the real world Gen1 Automotive Event Dataset for object detection, where we especially improve the detection of pedestrians by up to 5 \%.}

\onecolumn \maketitle \normalsize \setcounter{footnote}{0} \vfill

\section{\uppercase{Introduction}}
\label{sec:introduction}
\glsresetall
\Glspl*{dvs}, also known as event cameras, are vision sensors that register changes in intensity in an asynchronous manner. This allows them to record information with a much lower latency (in the range of milliseconds) compared to conventional RGB cameras. In addition, they have a very wide dynamic range (in the range of 140 dB) and can therefore even detect motion at night and in poor lighting conditions. Furthermore, \Gls*{dvs} cameras impress with their low energy consumption and they offer the possibility to filter out unimportant information (for example a stagnant background) in many applications. 

However, since this is a fairly recent technology, the mass of available data for deep learning approaches is very limited. Compared to RGB datasets like ImageNet \cite{ImageNet} with 14 million images, event datasets like N-CARS \cite{HATS} have only a few thousand labels. This makes data augmentation very important in order to avoid overfitting and to increase the robustness of the neural network. Even with larger datasets like the Gen1 Automotive Event Dataset \cite{gen1}, it has been shown that data augmentation significantly increases the performance of \Gls*{dl} methods, \cite{RVT}. 

Another challenge, particularly in the context of autonomous approaches, is the occurrence of occlusion, meaning that some objects are partially covered by other objects. In order to ensure save driving, \Gls*{dl} approaches have to be able to detect these occluded objects nevertheless. Regarding occlusion, current event augmentation methods only consider missing events either over time or in an area like it is done for RGB images. However, these approaches only model the behavior of an object moving in sync with the camera, which deviates from real-world scenarios. Additionally, automotive scenes are inherently dynamic and also event streams posses a temporal component. Consequently, the majority of objects within the scene are typically in motion relative to the camera, which can not be depicted by the simple dropping of events at a fixed location. 

Our objective is it to develop a more realistic augmentation technique, that accounts for the additional events generated by moving foreground objects. For this reason, we introduce \textbf{ShapeAug}, an occlusion augmentation approach, which simulates the events as well as the occlusions caused by objects moving in front of the camera. Our method utilizes a random number of objects and also a randomly generated linear movement in the foreground. We evaluate our augmentation technique on the most common event datasets for classification and further demonstrate its applicability in a real-world automotive task using the \textit{Gen1 Automotive Event Dataset} \cite{gen1}. Since \glspl{snn} \cite{gerstner_kistler_2002} share the asynchronous nature as well as the temporal component of event data, they are the natural choice for processing events and thus we choose to use \glspl{snn} for all our experiments.

Our contribution can be summarized trough the following points: We introduce \textbf{ShapeAug}, a novel occlusion augmentation method for event data, and assess its effectiveness for classification and object detection tasks. Furthermore, We evaluate the robustness of \textbf{ShapeAug} in comparison to other event augmentation techniques on challenging variants of the \textit{DVS-Gesture} \cite{dvsgesture} dataset.

\section{\uppercase{Related Work}}
\label{sec:relatedwork}

\subsection{Occlusion-aware RGB Image Augmentation}
Regarding RGB image augmentation, there are two main methods for statistical input-level occlusion: Hide-and-Seek \cite{hideandseek} and Cutout \cite{cutout}. Hide-and-Seek divides the image into $G \times G$ patches and removes (meaning zeros out) each patch with a certain probability. Cutout on the other hand chooses $N$ squares with a fixed side length and randomly chooses their center point in order to drop the underlying pixel values. The work of \cite{occ_img_cls} builds upon these methods while including a gradient-based saliency method as well as Batch Augmentation \cite{BatchAug}. These augmentations, however, are not able to mimic occlusion in real world event data, since event data has a temporal component meaning occluding objects would move and produce events themselves. 

\subsection{Event Data Augmentation}

Many data augmentation techniques for event data are adaptations of augmentation methods for RGB images. The work of \cite{NeuroDataAug} applies known geometric augmentations, including horizontal flipping, rolling, rotation, shear, Cutout \cite{cutout} as well as CutMix \cite{cutmix}. Geometric augmentations are a widely established technique, which is why we will use it in combination with our own augmentation method.

CutMix is a method to combine two samples with labels using linear interpolation. The EventMix \cite{eventmix} augmentation builds upon the idea of CutMix and applies it on event input data. However, this method is not able to realistically model occlusion, since it does not consider that the body of a foreground object may completely cover the background object. 

Inspired by Dropout \cite{dropout}, the authors of \cite{eventdrop} propose EventDrop, which drops events randomly, by time and by area. However, this method is not able to simulate occlusion in real-world dynamic scenes, since only objects moving in sync with the camera would not generate additional events themselves. Therefore, our method simulates not only the occlusion caused by foreground objects but also the resulting events and their own movement.  

\section{\uppercase{Method}}
\label{sec:method}

\subsection{Event Data Handling}

An event camera outputs an event of the form $e_i = (x_i, y_i, t_i, p_i)$, when the pixel at position $(x_i, y_i)$ and at time $t_i$ registers a logarithmic intensity change with a positive or negative polarity $p_i \in \{0,1\}$. Due to asynchronous nature of the camera, its output is very sparse and thus difficult to handle by neural networks. Therefore, we create event histograms $E$ in the shape $(T, 2, H, W)$ with $(H,W)$ as the height and width of the event sensor, where one event sample is split into $T$ time steps. We keep the polarities separated and feed the time steps consecutively into the network.
A set of events $\mathcal{E}$ is thus processed in the following way:
\begin{multline}
    \mathcal{E}(\tau, p, x,y) = \\ \sum_{e_i \in \mathcal{E}} \delta(\tau - \tau_i) \delta(p - p_i) \delta(x - x_i) \delta(y - y_i),
\end{multline}
\begin{equation}
    \tau_i =  \left \lfloor{\frac{t_i - t_a}{t_b -t_a}\cdot T}\right \rfloor,
\end{equation}
with $\delta(\cdot)$ as the Kronecker delta function.

\begin{figure}
     \centering
     \begin{subfigure}[b]{0.5\linewidth}
         \centering
         \includegraphics[width=\linewidth]{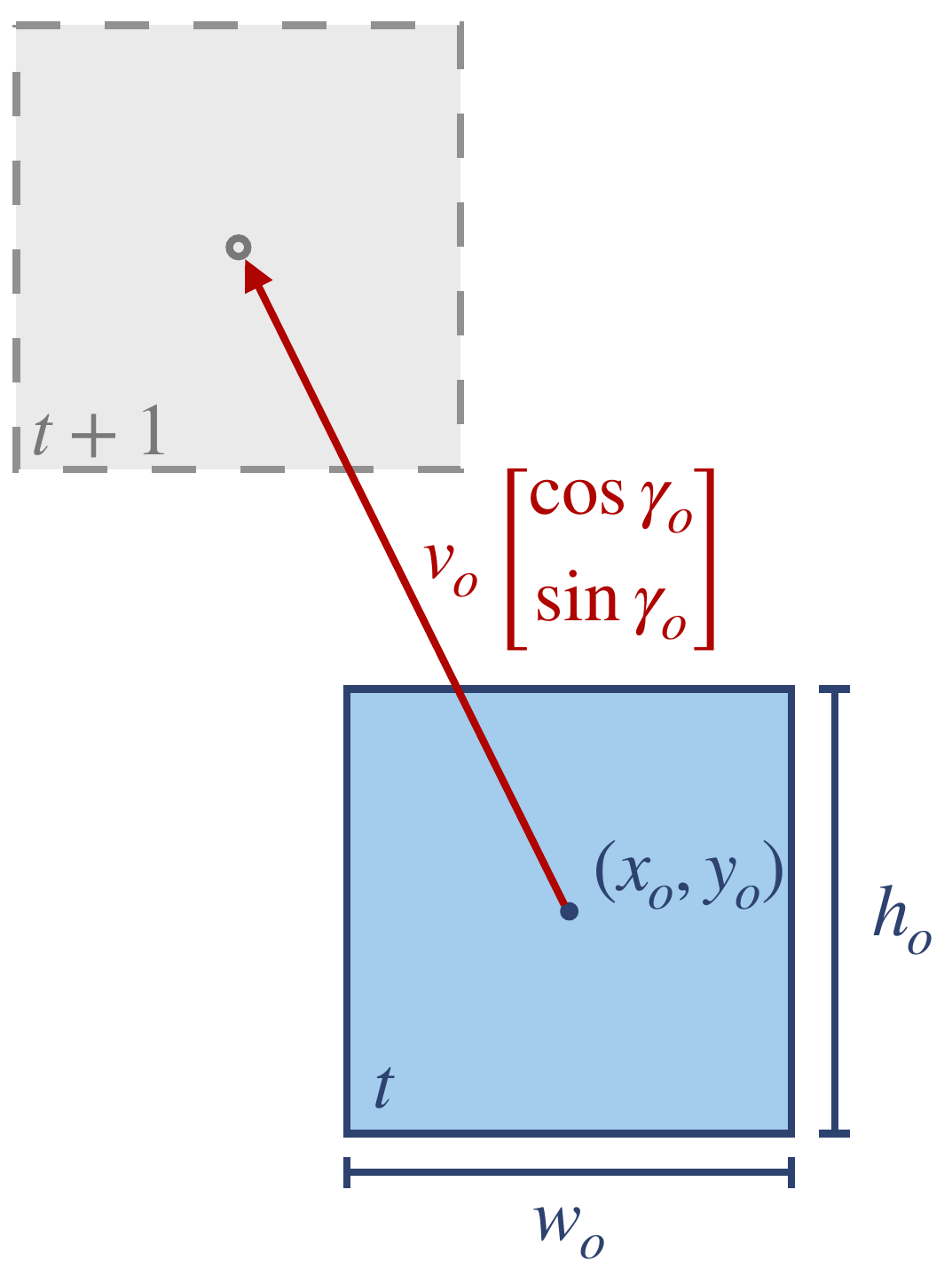}
         \caption{Shape movement.}
         \label{fig:shape1}
     \end{subfigure} \hspace{0.05\linewidth}
     \begin{subfigure}[b]{0.4\linewidth}
         \centering
         \includegraphics[width=\linewidth]{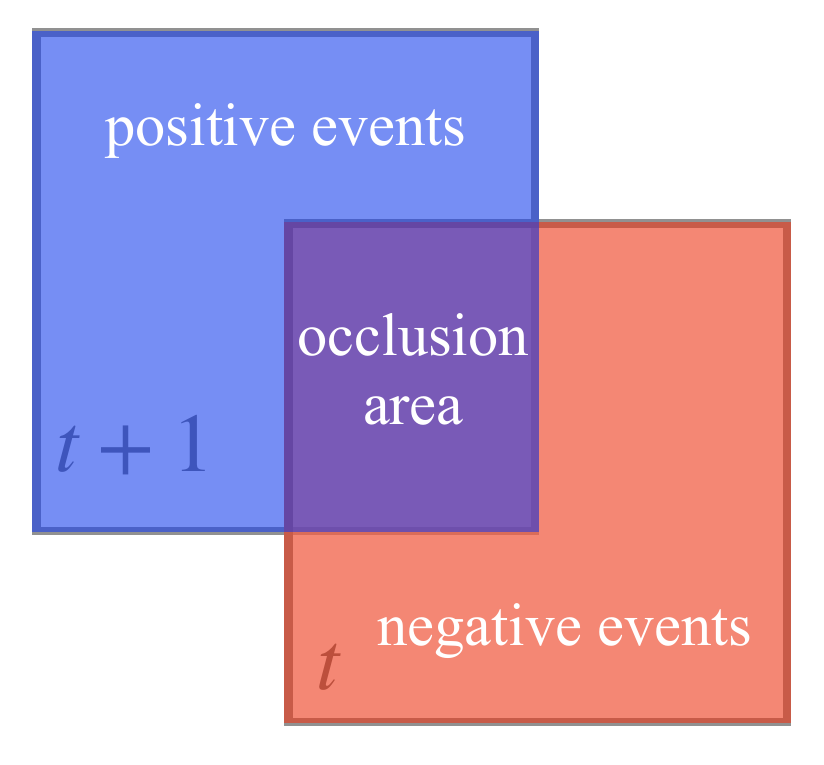}
         \caption{Positive and negative event computation.}
         \label{fig:shape2}
     \end{subfigure}
     \vspace{0.5em}
     \caption{Visualization of the shape parameters (position, size and direction) that are randomly chosen for each simulated object (\subref{fig:shape1}). The objects move between timesteps and are used to simulate the events that their movement would cause (\subref{fig:shape2}).}
      \label{fig:shapemovement}
\end{figure}

\subsection{Shape Augmentation}

Our occlusion-aware event augmentation is based on random objects moving on linear paths in the foreground. Since event streams have a temporal component, it is necessary to treat them similar to videos instead of image data. Thus it is important to avoid unnecessary noise corruptions of the temporal relations between time steps or frames. That is why we keep the augmentation consistent between time steps in an event stream, like the authors of \cite{Isobe2020} keep their augmentation consistent over all frames in a video sequence. To do so, we generate $N \in [1,5]$ random objects (circle, rectangle, ellipse), where each object $o$ gets assigned a random starting position $(x_o, y_o)$, a random size $h_o, w_o \in [3px, s_{max}]$ as well as a speed $v_o$ and an angle $\gamma_o$. The shape parameters are illustrated in \cref{fig:shape1}. We visualize these objects at their respective position and create frames for each timestep of the input event histogram. Between frames, the objects move linearly in the direction of a vector that is based on their assigned angle and speed:
\begin{equation}
    \begin{bmatrix}x^{t+1}_o\\ y^{t+1}_o\end{bmatrix} = \begin{bmatrix}x^{t}_o\\ y^{t}_o\end{bmatrix} + v_o \begin{bmatrix}\cos{\gamma_o}\\ \sin{\gamma_o}\end{bmatrix}
\end{equation}
with $t$ as the index of the frame. Whenever an object moves outside of the frame, a new object is created to maintain a consistent object count within the frame. 

In order to keep the needed computational overhead of the augmentation low, we choose a straight forward simulation technique for the events caused by the moving objects. Since real \glspl{dvs} register changes in the intensity, we simply use the difference between consecutive frames to find areas of events as illustrated in \cref{fig:shape2}. In these frames the background is colored black and the shapes are assigned the color gray. We further clip the frame difference to the mean of the non-zero event sample values in order to resemble the input more closely. If the difference at an image position is positive, it corresponds to a positive polarity event. Conversely, a negative difference indicates an event with negative polarity. However, since \glspl{dvs} exhibit a certain level of noise, we remove events with a probability of $p= 0.2$. 

\begin{figure*}
         \centering
         \includegraphics[width=0.79\linewidth]{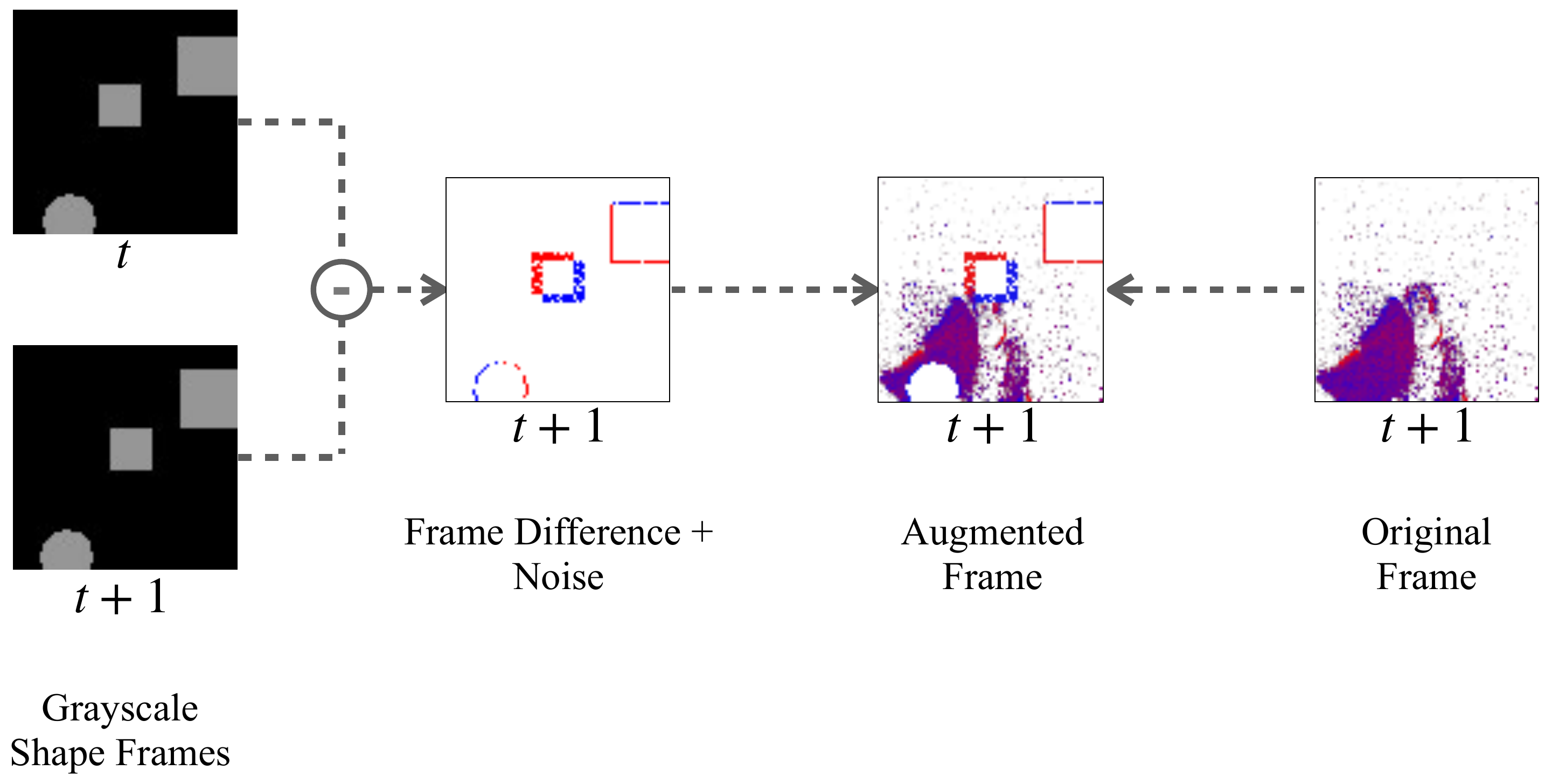}
         \caption{ShapeAug pipeline example for \textit{DVS-Gesture} \cite{dvsgesture}.}
         \label{fig:augpipeline}
\end{figure*}

We then include the generated events in the foreground of the original sample. However, since our goal is the modeling of occlusion, we remove the events in the sample that would be occluded by our simulated objects. This is because events are only triggered at the edge of moving shapes, which are homogeneously colored, not inside of them, where no intensity changes occur. The whole pipeline of our ShapeAug method is visualized in \cref{fig:augpipeline}.

\section{\uppercase{Experiments and Results}}
\label{sec:experiments}

\subsection{Datasets}
We validate our approach for classification on four \gls{dvs} datasets, including two simulated image-based datasets and two recordings of real movements with a \gls{dvs}. 
\textit{DVS-CIFAR10} \cite{cifar10dvs} is a converted \gls{dvs} version of the Cifar10 \cite{cifar10} dataset. It includes 10,000 event streams, which were recorded by smoothly moving the respective image in front of a \gls{dvs} with a resolution of $128px \times 128px$. \textit{N-Caltech101} \cite{ncaltech101} is likewise a converted dataset based on Caltech101 \cite{caltech101} containing 8709 images with varying sizes. \textit{N-CARS} \cite{HATS} is a real world \gls{dvs} dataset for vehicle classification. It contains 15422 training and 8607 test samples with a resolution of $120px \times 100px$. \textit{DVS-Gesture} \cite{dvsgesture} is a real world event dataset for gesture recognition. The dataset includes 11 hand gestures from 29 subjects resulting in 1342 samples with a size of $128px \times 128px$. For the datasets, which do not include a pre-defined train-validation split definition, we used the same split as \cite{eventmix}. We further resize all the event streams to a resolution of $80px \times 80px$ using bi-linear interpolation, before applying any augmentation and divide them into 10 timesteps. 

For the object detection task we choose the \textit{Gen1 Automotive Event Dataset} \cite{gen1}, which is recorded by a \gls{dvs} with a resolution of $304px \times 240px$ during diverse driving scenarios. It includes 255k labels with bounding boxes for pedestrians and cars. Like previous work \cite{ASTMNet},  \cite{RVT}, \cite{1Megapixel}, we remove bounding boxes with a diagonal less than $30px$ or a width or height less than $10px$. We create discretized samples with a window size of 125ms and further divide them into 5 timesteps.

\subsection{Implementation}

We choose \glspl{snn} for conducting all our experiments, since their asynchronous and temporal nature fits well to the processing of event data. Furthermore, \glspl{snn} are inherently more energy efficient than comparable \glspl{ann}, making them particularly useful for real world automotive applications.

For our classification experiments, we adopt the training implementation of \cite{eventmix} and therefore use a preactivated ResNet34 \cite{PreActResnet} in combination with PLIF neurons \cite{PLIF}. As an optimizer we use AdamW \cite{AdamW} with a learning rate of 0.000156 and a weight decay of $1\times 10^{-4}$. We further apply a cosine decay to the learning rate for 200 epochs. Because it is already an established practice, we apply geometrical augmentation on all our experiments including the runs for our baseline. Similar to \cite{eventmix}, we therefore apply random cropping to size $80px \times 80px$ after padding with $7px$, as well as random horizontal flipping and random rotation by up to $15^{\circ}$.

Regarding our object detection experiments, we follow the example of \cite{ODSNN} and choose a Spiking DenseNet \cite{Huang2017} with SSD \cite{SSD} heads as our network. We also use an AdamW optimizer \cite{AdamW} with a learning rate of $5\times 10^{-4}$ and a cosine decay for $100$ epochs. Moreover, we utilize a weight decay of \num{1e-4} as well as a batch size of $64$. As in \cite{ODSNN}, we apply a smooth L1 loss for the box regression and the Focal Loss \cite{Focalloss} for the classification task. Following the example of \cite{RVT}, we utilize the following geometrical augmentations for all experiments: Random zoom-in and -out as well as random horizontal flipping. Since rotations distort bounding boxes, we choose to not apply this augmentation method for the object detection task. As a metric for our results, we use the mean Average Precision (mAP) over 10 IoU thresholds [.5:.05:.95].

\subsection{Event Data Classification}

\begin{table*}
\caption{Comparison of classification results using different max shape sizes $s_{max}$. We report the top-1 accuracy as well as the top-5 accuracy in brackets, except for datasets with less than 5 classes. The best and second best results are shown in \textbf{bold} and \underline{underlined} respectively. }
\label{tab:classification}
\begin{center}
\begin{tabular}{cccccc}
 \hline 
Method  & Max Shapesize [px] &DVS-CIFAR10& N-Caltech101 & N-CARS&  DVS-Gesture \\
 \hline 
 Geo  & - & 73.8 (95.5) & 62.2 (81.5) & \underline{97.1} &89.8 (99.6)\\
 Geo + ShapeAug  & 10 & \underline{74.3} (95.1) & 68.0 (83.9)& \textbf{97.3} & \underline{90.9} (99.6)\\
 Geo + ShapeAug  & 30 & 73.9 (94.7)& \textbf{68.7} (86.9) & 96.9 &\textbf{91.7} (100)\\
 Geo + ShapeAug & 50 &\textbf{75.7} (96.7)& \underline{68.2} (85.2)& 96.9 &90.5 (99.2)\\
 \hline
\end{tabular}
\end{center}
\end{table*}

\autoref{tab:classification} shows the results of our augmentation technique on various event classification datasets using different maximum shape sizes $s_{max}$. Our method is able to outperform the baseline for all four datasets. Especially the data, that was recorded based on RGB images, benefited greatly from the shape augmentation. This may be caused by the similarity of movements, between the recording and the simulated shapes. More dynamic scenes may require more complex movements of the shapes, which however can lead to an increased simulation overhead. Additionally, it is challenging to further improve the results on the N-CARS dataset, since the baseline is already able to nearly perfectly classify the validation set. The results also show that, in the majority of cases, even using $s_{max} = 50$ improves the results. However, the best choice of shape sizes depends on the actual objects depicted in the dataset.

\subsection{Comparison with Existing Literature on Robustness}

\begin{table*}[t]
\caption{Comparison of the robustness of current event augmentation approaches on various augmented versions of DVS Gesture \cite{dvsgesture}. The best and second best results are shown in \textbf{bold} and \underline{underlined} respectively.}
\label{tab:robustness}
\begin{center}
\begin{tabular}{cccccccc}
 \hline 
\backslashbox{Train}{Valid}  &- & Geo & Drop  &  Shape  \\
 \hline 
 Geo &  89.8 & 87.5 & 58.0 &  63.6\\
 Geo + Drop & 89.8 & 87.9 & 86.0& 73.9  \\
 Geo + Mix & 93.1 & \underline{92.4} & 82.6  &  76.9 \\
 Geo + Shape & 91.7 & 90.5 & 84.1 & 87.9 \\
 Geo + Drop + Shape & 91.7 & 90.2 & 88.6 & 87.5\\ 
 Geo + Mix + Shape & \underline{94.7}&  91.7 &  87.5 & \textbf{91.3}\\
 Geo + Mix + Drop & 92.8 & 89.8 & \underline{89.8} & 75.4\\
 Geo + Mix + Drop + Shape & \textbf{95.8} & \textbf{94.7} & \textbf{92.8} & \underline{89.8}\\
 \hline
\end{tabular}
\end{center}
\end{table*}

In order to evaluate and compare the robustness of our event augmentation, we create three challenging validation datasets based on DVS Gesture \cite{dvsgesture} using the following augmentations on each sample: Geometric (horizontal flipping, rotation, cropping), EventDrop \cite{eventdrop} and ShapeAug with $s_{max}=30$. Notably, we decided to not include the EventMix \cite{eventmix} for validation augmentation due to its generation of multi-label samples, which would create an unfair comparison, as the other methods were not trained for that case. The results of our experiments can be found in \autoref{tab:robustness}.

Since augmentation techniques have the unique property that they can be combined, we decided to not only compare the existing approaches but also investigate their combination during training. Generally we applied every augmentation with a probability of $p = 0.5$, resulting in some training examples undergoing multiple augmentations. Geometrical augmentation was also here universally applied to all training samples.

\paragraph{Comparison to State-of-the-Art.} 
The results on the standard validation dataset show, that ShapeAug is able to improve the networks performance much more than EventDrop, which approximately achieves the same accuracy like the baseline. Generally, EventMix has the greatest positive impact on the outcome. Nevertheless, it has to be noted that the mixing of samples, allows the network to see every sample multiple times during an epoch which is not the case for other augmentation techniques. 

\paragraph{Robustness.}
Also on most of the augmented validation data, ShapeAug is able to outperform EventDrop, proving its increased robustness. Only on the drop-augmented data, EventDrop naturally performs better in comparison to ShapeAug and EventMix. Furthermore, no method that was trained without ShapeAug is able to perform well on the shape-augmented data ($>10 \%$ lower accuracy than ShapeAug), showing their lack of robustness against moving foreground objects in event data. Conversely, methods trained with ShapeAug or EventMix are still capable of predicting drop-augmented data (only $2-4 \%$ lower accuracy compared to EventDrop), which proves their natural robustness against this augmentation. 

\paragraph{Combination of Methods.}
Our ShapeAug method shows great potential for being combined with other augmentation techniques, since it is always able to improve the performance on the standard as well as all the challenging validation data. The usage of EventDrop on the other hand even leads to a decreased accuracy when combined with multiple approaches on most of the validation data. The best performance, outperforming the baseline by about $6 \%$, can be achieved when we train on all event augmentation techniques combined during training. Overall, however, we find that our ShapeAug augmentation can nearly compensate for all the benefits of utilizing drop-augmentation.

\subsection{Automotive Object Detection}

\begin{table*}
\centering
\caption{Object detection test results on the \textit{Gen1 Automotive Event Dataset} \cite{gen1}.}
\label{tab:gen1_od}
\begin{tabular}{ccccccc}
\hline 
 Method & Max Shape size [px]& mAP & AP$_{50}$ & AP$_{75}$ & AP$_{car}$ & AP$_{ped}$\\
 \hline
Geo & - & 29.26 & 55.20 & 26.99 & 41.13 & 17.37\Tstrut\\
Geo + ShapeAug & 50 & 29.60 & 56.99 & 26.70 & 40.91 & 18.30\\
Geo + ShapeAug & 100 & 27.43 & 52.16 & 25.59 & 40.09 & 14.78\\
Geo + ShapeAug & 150 & 26.33 & 50.59 & 24.36 & 39.43 & 13.24
\\
\hline 
\end{tabular}
\end{table*}

\begin{table*}
\centering
\caption{Evaluation of robustness over multiple augmented test sets of the \textit{Gen1 Automotive Event Dataset} \cite{gen1}.}
\label{tab:gen1_od_robustness}
\begin{tabular}{cccccc}
\hline 
 Method & Max Shape size [px]& - & Geo &Drop & Shape \\
 \hline
Geo & - &  29.26 & 29.15 & 23.48 & 25.68\\
Geo + ShapeAug & 50 & 29.60 & 28.76 & 24.21 & 27.94\\

\hline 
\end{tabular}
\end{table*}

We further test our ShapeAug technique for object detection on the \textit{Gen1 Automotive Event Dataset} \cite{gen1}. The results can be seen in \autoref{tab:gen1_od} and show that ShapeAug is able to increase the performance of the detection. It especially has a positive impact on the bounding box prediction of pedestrians, where it increases the AP by over $5 \%$. Pedestrians are in general much more challenging to detect, since they usually appear smaller than cars in the images and their movements as well as their appearances are much more complex and have a high variance. Furthermore, they are more prone to be occluded by other traffic participants, which may be the reason ShapeAug is especially benefiting their detection. However, the results also indicate that the size of the shapes has a significant impact on the predictions. Compared to the classification dataset, the objects in the \textit{Gen1 Automotive Event Dataset} can appear very small and too much occlusion may hinder the training signal to pass trough the network effectively. Furthermore, the movements of objects in real world automotive scenes are very complex and do not just follow a linear pattern. Therefore, the occlusion augmentation can be further improved by increasing the complexity of movements and shapes, which however will also increase the simulation overhead. 

\paragraph{Robustness Analysis.} 

Also for the object detection task, we examined the robustness of ShapeAug on different augmented test sets of the \textit{Gen1 Automotive Event Dataset} \cite{gen1}. Regarding the geometrical augmentation, we applied zoom (either zoom-in or zoom-out) on all images as well as horizontal flipping with a probability of $p=0.5$. Since our experiments were done on pre-processed event data in order to decrease the computations during training, we opted for Random Erasing \cite{randomerase}, which randomly erases rectangles of the input image, as our drop augmentation. \autoref{tab:gen1_od_robustness} contains the results of our robustness evaluation. If the appropriate size of shapes is chosen, ShapeAug indeed improves the results for the prediction on shape-augmented as well as on drop-augmented data. This means for tasks where a high degree of occlusion during inference is expected, ShapeAug can be a valuable technique to increase prediction performance. However, it is necessary to evaluate the right values for hyperparameters, including the shape size, the number of objects as well as  the movement pattern of the shapes. 

\section{\uppercase{Conclusion} }
\label{sec:discussion}

Augmentation for event data during the training of neural networks is crucial in order to ensure robustness as well as to avoid overfitting and to improve accuracy. In this work, we introduced ShapeAug, an augmentation technique simulating moving foreground objects in event data. Our method includes the simulation of a random amount of objects, moving on randomly chosen linear paths, and using the resulting events from their movement. Since the objects are in the foreground, ShapeAug allows the modeling of realistic occlusions, since in real world scenarios occluding objects would cause the generation of events. 

We show the effectiveness of our approach on the most common event classification datasets, where it is able to improve the accuracy significantly. Furthermore, ShapeAug proves to increase the robustness of predictions on a set of challenging validation data and is able to outperform other event drop augmentations. Our technique can be also easily combined with other augmentation methods, leading to an even higher boost of the prediction performance. Additionally, ShapeAug improved the object detection on a real world automotive dataset and further enhanced the robustness against various augmentations on the test dataset.

Currently, our shape augmentation method only simulates very simple homogenously colored shapes and their movements. It remains for future work to explore the simulation of complex movements and more sophisticated textures and object shapes.

\section*{\uppercase{Acknowledgements}}
This work was funded by the Carl Zeiss Stiftung, Germany under the Sustainable Embedded AI project (P2021-02-009) and partially funded by the Federal Ministry of Education and Research Germany under the project DECODE (01IW21001).

\bibliographystyle{apalike}
{\small
\bibliography{paper}}

\end{document}